\documentclass[conference]{IEEEtran}
\IEEEoverridecommandlockouts
\def\IEEEtitletopspace{18pt}
\usepackage{cite}
\usepackage{amsmath,amssymb,amsfonts}
\usepackage{algorithmic}
\usepackage{graphicx}
\usepackage{flushend}
\usepackage[capitalise]{cleveref}
\usepackage{textcomp}
\usepackage{xcolor,colortbl}
\usepackage[top=54pt, left=54pt, right=54pt, bottom=54pt]{geometry}%
\usepackage{url} 
\usepackage{graphicx}
\usepackage{makecell}
\usepackage[table]{xcolor}
\usepackage{siunitx}
\usepackage{subcaption}

\usepackage{tabularx}
\usepackage{booktabs}
\usepackage{threeparttable}

\definecolor{Gray}{gray}{0.90}

\def\BibTeX{{\rm B\kern-.05em{\sc i\kern-.025em b}\kern-.08em
    T\kern-.1667em\lower.2ex\hbox{E}\kern-.125emX}}
\begin{document}

\definecolor{Gray}{gray}{0.75}
\definecolor{Grey}{gray}{0.92}

\newcolumntype{a}{>{\columncolor{Grey}}c}
\title{\LARGE \bf MOSS: Multimodal Origami Swamp Robot System  }



\author{ Tian Lan$^{1,2}$, Atar Babgei$^{1,2}$, Luca Romanello$^{2,3}$, Mirko Kovac$^{2,3}$, Sophie F. Armanini$^{1}$ \\

\thanks{

 \par $^{1}$Aerial Robotics Laboratory, Imperial College London.
 \par $^{2}$Laboratory of Sustainability Robotics, EMPA, Dübendorf, Switzerland.
 \par $^{3}$Laboratory of Sustainability Robotics, EPFL, Lausanne Switzerland.
}}

\maketitle
\begin{abstract}

Robotic systems for environmental monitoring and ecological surveying in wetlands and swamps must operate across diverse terrestrial and aquatic conditions. Conventional fixed-geometry wheels cannot adapt their contact geometry to these varying conditions. This work presents MOSS (Multimodal Origami Swamp Robot System), a compact amphibious robotic platform with 3D-printed origami wheels. The compact crank mechanism switches the wheel between contracted and expanded configurations, changing its width, body diameter, and fin orientation simultaneously. The contracted configuration is better suited to efficient rolling on firmer ground, whereas the expanded configuration provides greater contact and terrain engagement on softer surfaces. MOSS was evaluated on soil, sand, mud with varying water content, and water, in addition to in-motion morphing and terrestrial-aquatic transition tests. The more energy-efficient wheel configuration depended on terrain. The contracted configuration achieved a lower cost of transport on firmer soil and mud, whereas the expanded configuration performed better on wetter mud, sand, and water, reducing the cost of transport by up to 29\%. Expanded wheels also reduced slip in wet mud and reduced both slip and hub sinkage in sand. MOSS successfully completed configuration changes in \textbf{1.95\,\textpm\,0.11~s} and demonstrated continuous locomotion across soil–mud–water transitions. Outdoor trials further demonstrated operation across natural deformable terrain and water. These results show that adaptive wheel geometry provides terrain-dependent locomotion benefits and enables a single compact platform to traverse substantially different terrestrial and aquatic conditions.

\end{abstract}


\section{Introduction}
\label{sec:introduction}

Environmental change and habitat loss motivate long-term, low-disturbance ecological monitoring for conservation and ecosystem management~\cite{reid2005ecosystems,pimm2014biodiversity}. Swamps support diverse communities, regulate water, and store carbon~\cite{worldrainforests_swamps}. Yet biodiversity observations remain spatially and temporally uneven, while seasonal changes in water level and soil condition make repeated surveys difficult~\cite{robertson2025global}. Robotic platforms can reduce the need for repeated human access while collecting observations in challenging environments~\cite{dunbabin2012robots,lowman2021life,bujari2020mobile,siddall2014launching,debruyn2020medusa}.

\begin{figure}[htb!]
    \centering
    \includegraphics[width=1\linewidth]{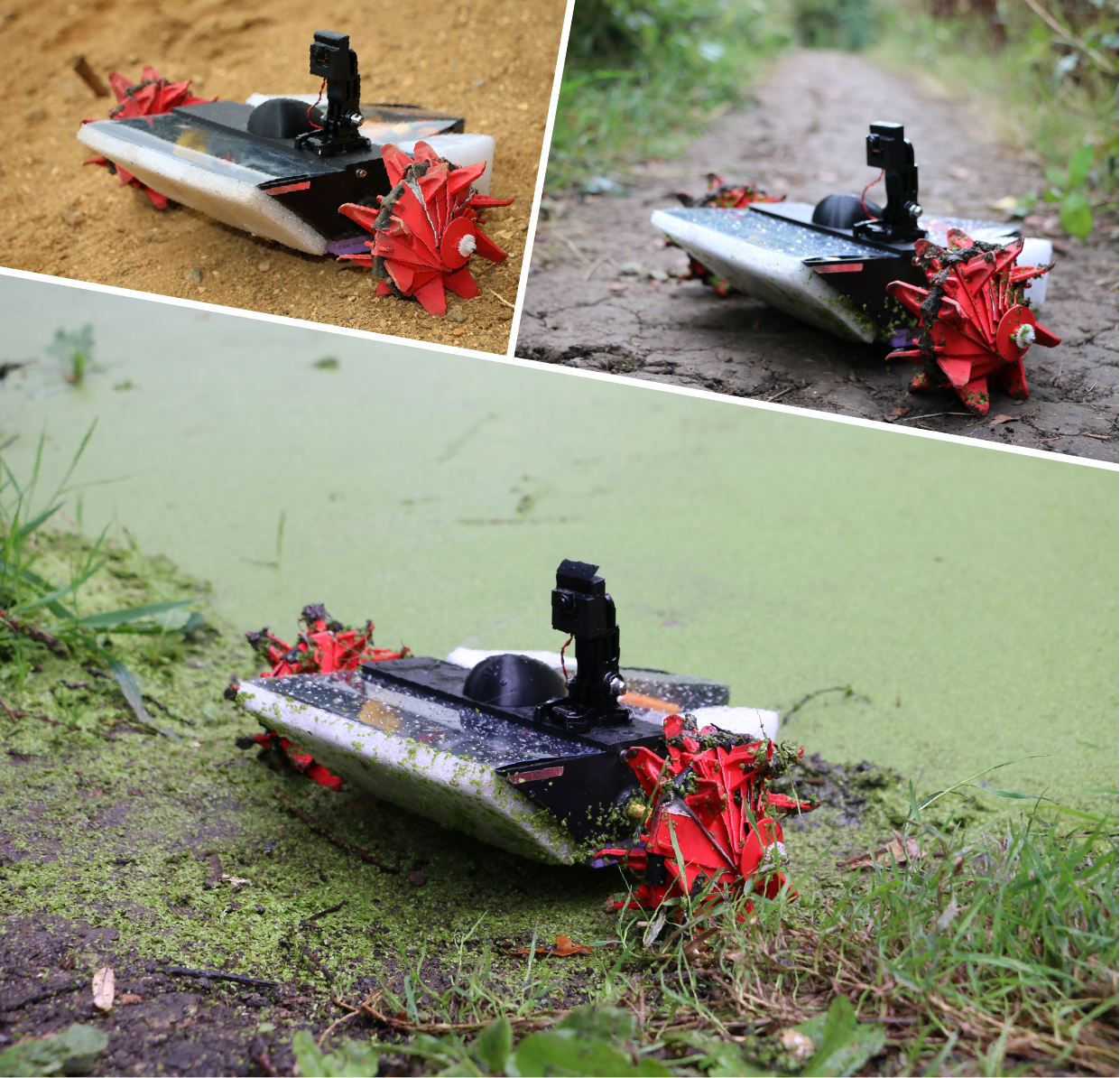}  
    \caption{MOSS operating on representative swamp-relevant terrain}
    \label{fig:system}
\end{figure}

However, swamp environments often combine firm ground, sand, soil, mud, and water. These surfaces differ in how they support the robot and provide traction, and soft ground can also lead to wheel sinkage~\cite{zhou2017multimodal}. Firm ground favours low rolling resistance, whereas deformable terrain requires sufficient contact area and shear engagement to limit slip and sinkage. Fixed-geometry wheels cannot change their diameter, width, or contact geometry as these requirements change. Previous studies have shown that transformable and origami-inspired wheels can reconfigure their geometry to adjust ground contact across different terrains~\cite{kim2014wheel,lee2013deformable,lee2017origami,rhoads2016design,lee2024variable}. Although these studies establish the feasibility of wheel morphing, integrating axial reconfiguration, terrain-interacting fins, and drive-torque transmission into a single compact mechanism remains challenging. More importantly, the locomotion performance of such multimodal morphing wheels remains insufficiently characterised across swamp-relevant terrestrial and aquatic conditions. To address these gaps, this work presents MOSS (Multimodal Origami Swamp Robot System), an origami wheel mobile platform developed for locomotion across swamp-relevant deformable terrains. With MOSS, a systematic experimental evaluation is conducted across representative terrestrial and aquatic environments to determine how morphing between contracted and expanded configurations alters wheel–surface interactions and delivers terrain-specific locomotion benefits. 

The main contributions of this work are:

\begin{itemize}

    \item Design and fabrication of a 3D-printed origami morphing wheel with integrated fins for wetland locomotion.
    
    \item Integration of the morphing wheels into a compact amphibious robot, using telescopic drive shafts that transmit drive torque while enabling axial reconfiguration.

    \item Laboratory tests showing contracted wheels are more efficient on firmer ground and expanded wheels on softer ground and water, plus outdoor demonstrations.
    
    
\end{itemize}


\section{Morphing Wheel Design and Manufacturing}
\label{sec:wheel}

\subsection{The Origami Wheel}
\label{subsec:wheel_fabrication}

\begin{figure}[t!]
    \centering
    \includegraphics[width=\linewidth]{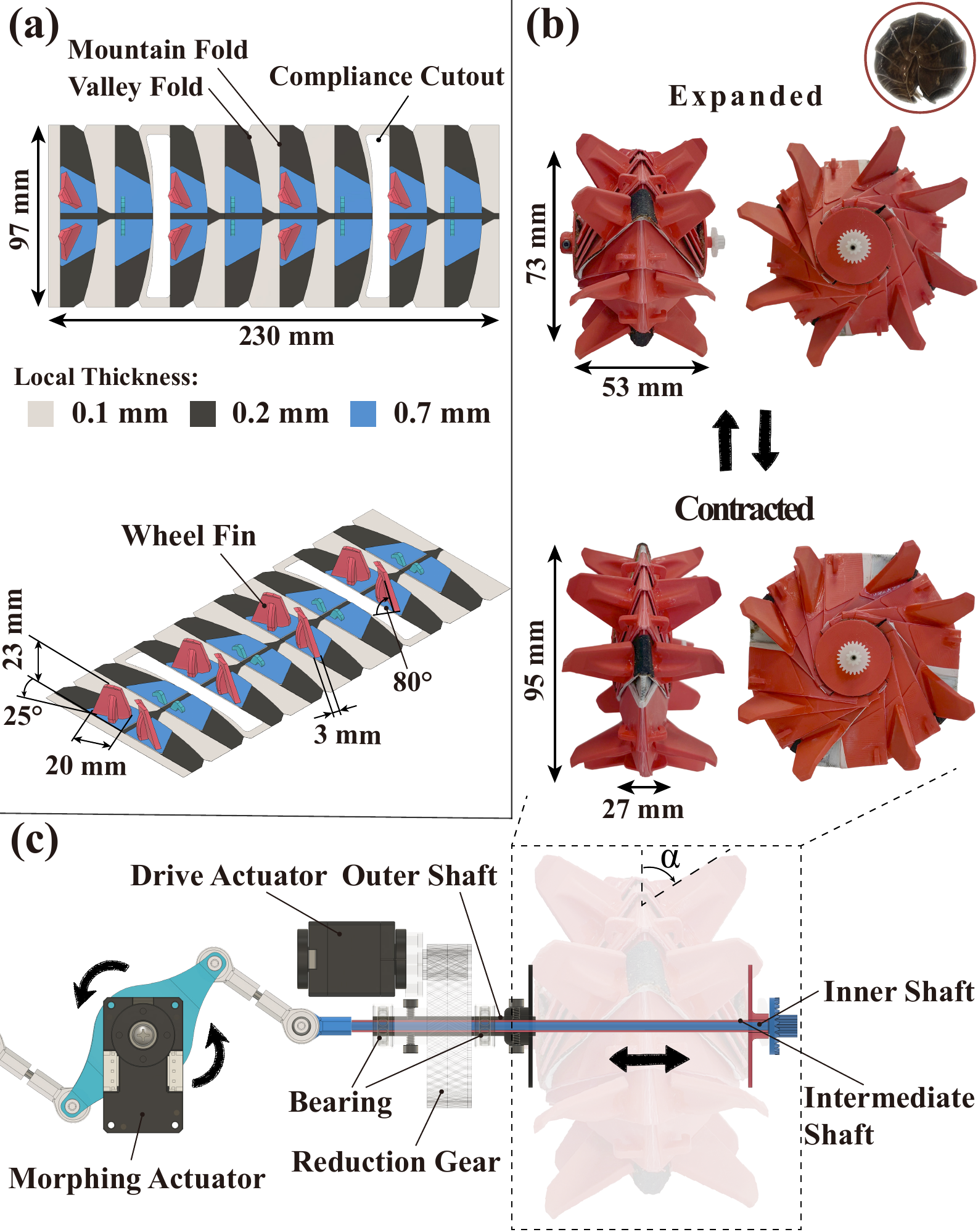} 
    
    \caption{Design and characterisation of the origami wheel. (a) The modified origami pattern used in this work for 3D printing, showing the local thickness distribution and integrated fins. Each wheel is formed by bonding two identical printed pieces into a ring and folding it. (b) Wheel dimensions in the expanded and contracted configurations. (c) Operating principles of the morphing and driving mechanisms and the telescopic-shaft design. (Pillbug photograph \cite{lough2026rolling})}
    \label{fig:Actuation_vertical}
\end{figure}

The wheel was designed to change the axial width, body diameter, and terrain-contact geometry through a single morphing motion. Its folding pattern was adapted from the rotational-sweep Origami Sphere proposed by Mitani~\cite{mitani_design_2009}, but was modified for thick-panel additive manufacturing. The resulting origami pattern is shown in Fig.\ref{fig:Actuation_vertical} (a). Each wheel is made from two identical 3D-printed patterns bonded into a ring. The ring is then folded by hand along the designed crease lines to form the wheel structure. Unlike an ideal paper model, a printed wheel must accommodate finite panel thickness, crease interference, limited folding clearance, and repeated contact loading~\cite{chen2015origami,callens2018flat}. The pattern therefore uses three local thicknesses of $0.1$, $0.2$, and $0.7~\mathrm{mm}$, as indicated by the colours in Fig.~\ref{fig:Actuation_vertical}(a). The thinner regions provide folding compliance, whereas the thicker panels provide the load-bearing and wear-resistant structure. The central crease region is widened to reduce interference during deformation.

Compared with conventional multilayer fabrication \cite{rhoads2016design,lee2017origami}, printing each flat pattern as a single piece reduces the number of parts and avoids cutting, stacking, and aligning multiple layers, as well as many assembly steps. Digital fabrication allows local thicknesses, crease features, and geometry to be modified directly, accelerating design iteration. The resulting patterns improve geometric consistency between wheels and reduce assembly errors while preserving the intended folding sequence and final wheel configuration.

However, PLA has limited ductility. Thinning the folding regions improves their flexibility but reduces their load-bearing capacity and durability during repeated deformation. Compliance cutouts (Fig.~\ref{fig:Actuation_vertical}a) were therefore introduced in selected regions and bridged with soft textile tape. The tape provides local flexibility, while overlapping segments arranged like pillbug plates (Fig.~\ref{fig:Actuation_vertical}b) preserve the wheel's circumferential load path \cite{lough2026rolling}. The same tape reinforces the exposed segment tips and reduces wear during ground contact. Together, these features support repeated morphing while retaining the stiffness required for locomotion. 

The resulting transformation couples wheel width and body diameter. In the prototype, the contracted configuration is approximately $27~\mathrm{mm}$ wide with a $95~\mathrm{mm}$ body diameter, whereas the expanded configuration is approximately $53~\mathrm{mm}$ wide with a $73~\mathrm{mm}$ body diameter. The wheel width nearly doubles during reconfiguration. These dimensions describe the wheel body without the fins and were selected based on the available chassis envelope, actuator stroke, linkage packaging, and manufacturing constraints. 


\subsection{Integrated Fins and Coupled Morphing}
\label{subsec:wheel_fins}

As shown in (Fig.~\ref{fig:Actuation_vertical}a), fin-like features inspired by mudskipper pectoral appendages were integrated directly into the printed pattern~\cite{pace2009mudskipper}. During contact, the fins penetrate deformable terrain, increase tangential shear resistance, and distribute load over additional contact points. This design increases available traction and helps reduce local sinkage. The fins therefore form part of the wheel structure rather than a separately actuated attachment. Their orientation changes passively as the inclined origami panels rotate. In the contracted configuration, the fin edges are approximately parallel to the ground, corresponding to $\alpha\approx85^{\circ}$ under the geometric convention used here. As the wheel expands, $\alpha$ decreases below $85^{\circ}$, which may improve shear engagement and load distribution on deformable terrain such as sand and mud. The fins are considered an integral part of the morphing wheel, so their individual contribution is not isolated. 


\subsection{Crank Actuation and Shape Retention}
\label{subsec:wheel_actuation}

The wheel is morphed by a centrally mounted actuator and crank linkage (Fig.~\ref{fig:Actuation_vertical}c). The inner wheel sides remain fixed to the axle, while the linkage pulls the outer sides inward or pushes them outward. This changes the wheel's axial width and body diameter without changing the wheel-axle interface. The contracted and expanded configurations are defined by target positions of the morphing actuator and are used as the two locomotion conditions in this study. In position mode, the actuator actively maintains the selected wheel configuration without a separate mechanical lock.

\section{Telescopic Shaft and Robot Integration}
\label{sec:shaft_robot}

\subsection{Telescopic Torque and Axial Transmission}
\label{subsec:telescopic_shaft}

The morphing wheel imposes a coupled transmission requirement: the wheel must change its axial position while continuing to transmit drive torque under transverse wheel loading. An external hub or frame can provide a rigid load path but  increase the required stroke volume and reduce platform compactness~\cite{liu_origami-wheeled_2025}. Lee~\textit{et al.} used a coil-spring shaft with a central wire to transmit torsion while allowing axial compliance~\cite{lee2017origami}. However, the spring provided limited lateral constraint for the outer hub, which could cause wheel-axle misalignment in the expanded configuration and required additional hub support under load. Tether-driven mechanisms reduce rigid hardware but require careful routing and pretensioning. Moreover, their tension-only action requires an opposing cable, a second actuator, or an elastic element for the return stroke~\cite{moger_design_2024,murphy2024evaluation}.

Therefore, this work developed a telescopic shaft that separates axial translation for wheel morphing from rotation for locomotion (Fig.~\ref{fig:Actuation_vertical}c). A rotating outer shaft transmits drive torque to the wheel. The morphing actuator is fixed to the chassis and drives an inner shaft axially without co-rotating with the wheel. A brass intermediate shaft surrounds the carbon-fibre inner shaft and provides a guided lateral support. This arrangement shortens the unsupported span of the slender inner shaft for wheel morphing while preserving the axial motion required by the crank linkage.

The shaft layout was evaluated based on a first-order cantilever model. In the maximum-expanded configuration, the carbon-fibre inner shaft has a diameter $d=2~\mathrm{mm}$ and a maximum unsupported length of $L_0=63~\mathrm{mm}$. Its second moment of area is
\begin{equation}
I_i=\frac{\pi d^4}{64}=0.785~\mathrm{mm^4}.
\end{equation}
For a transverse wheel load $F_{\perp}$, the corresponding tip deflection is approximated by
\begin{equation}
\delta_0=\frac{F_{\perp}L_0^3}{3E_c I_i},
\end{equation}
where $E_c$ is the longitudinal flexural modulus of the carbon-fibre pultrusion. Assuming that the $560~\mathrm{g}$ robot mass is shared equally between the two driven wheels gives a nominal per-wheel load of $F_{\perp}\approx2.75~\mathrm{N}$.

Using the $20$--$30~\mathrm{GPa}$ flexural modulus range specified in the manufacturer's datasheet~\cite{easycomposites_carbon_fibre_pultrusions_2024}, the model predicts a $9.7$--$14.6~\mathrm{mm}$ tip deflection at full extension, indicating limited lateral stiffness of the unsupported inner carbon fibre shaft.

The intermediate shaft has outer and inner diameters of $3~\mathrm{mm}$ and $2~\mathrm{mm}$, respectively. Its geometric second moment of area is
\begin{equation}
I_m=\frac{\pi(D_o^4-D_i^4)}{64}=3.19~\mathrm{mm^4},
\end{equation}
which is approximately $4$ times $I_i$. This ratio represents only the geometric bending resistance and does not directly measure the stiffness of the assembled shaft. The intermediate shaft creates a shorter and more constrained load path around the inner carbon-fibre shaft. The model is therefore a first-order design estimate, not a complete strength validation.


\subsection{Robot Platform and Control}
\label{subsec:robot_integration}

\begin{figure}[hbt!]
    \centering
    \includegraphics[width=1\linewidth]{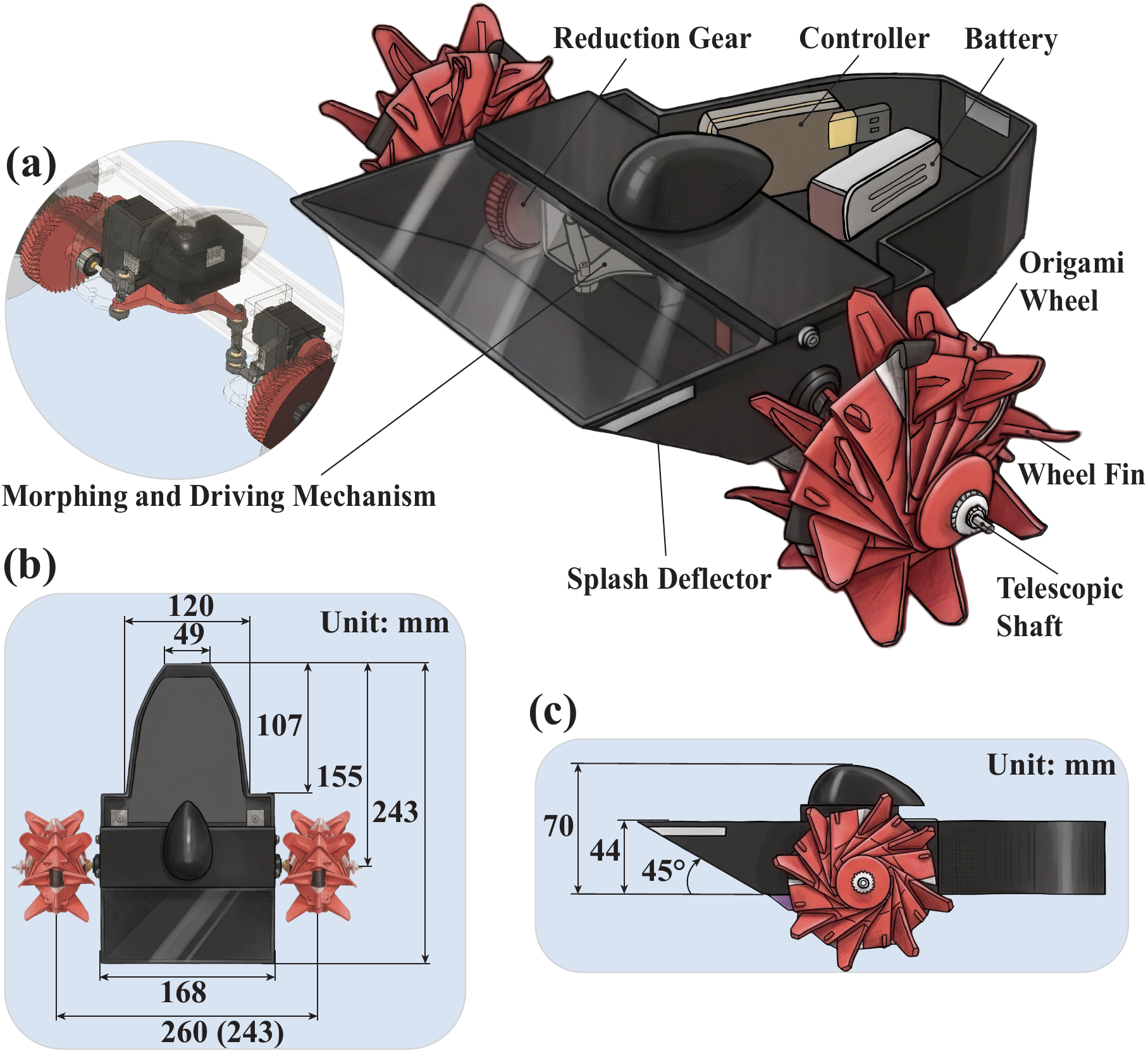} 
    
    \caption{Overall design of MOSS. (a) Morphing and wheel-driving mechanisms. A centrally mounted servo actuates a crank linkage that pulls or pushes the sides to switch between the contracted and expanded configurations.  (b,c) Dimensions of the robot. The wheel-to-wheel span: 243 mm (the contracted configuration) and 260 mm (expanded configuration).}
    \label{fig:SystemOverview}
\end{figure}

\begin{table}[hbt!]
  \centering
  \begin{threeparttable}
  \caption{Key hardware specifications of the robot.}
  \label{tab:robot_specs}
  \renewcommand{\arraystretch}{0.95}
  \setlength{\tabcolsep}{3pt}
  \small
  \begin{tabularx}{0.92\columnwidth}{
      @{}>{\raggedright\arraybackslash}p{0.34\columnwidth}
      >{\raggedright\arraybackslash}X@{}}
  \toprule
  \textbf{Specification} & \textbf{Value} \\
  \midrule
  Total mass           & 560\,g \\
  Origami wheel mass   & 40\,g each \\
  Drive actuators      & 2$\times$ Dynamixel XL330-M077-T \\
  Drive gear reduction & 6:1 \\
  Morphing actuator    & Dynamixel XC330-M288-T \\
  Battery              & 3S 850\,mAh LiPo \\
  Onboard computer     & Raspberry Pi CM4 (ROS 2 Humble) \\
  Sensors              & USB camera and microphone \\
  \bottomrule
  \end{tabularx}
  \end{threeparttable}
\end{table}

Table~\ref{tab:robot_specs} summarizes the main hardware components of the platform. The robot integrates two wheel-drive actuators, a central morphing actuator, a wedge-shaped chassis, and a modular electronics bay (Fig.~\ref{fig:SystemOverview}). Two Dynamixel XL330-M077-T actuators independently drive the left and right wheels through external $6{:}1$ herringbone reductions. The offset gears leave space for the telescopic shaft and increase the available wheel torque. A Dynamixel XC330-M288-T actuator drives the morphing crank in position mode and actively maintains either the contracted or expanded configuration without a mechanical lock.

The two drive actuators control the left- and right-wheel angular velocities, $\omega_L$ and $\omega_R$. Together with the active wheel configuration, these actuator states provide the inputs to the kinematic model in Sec.~\ref{subsec:kinematics}. The chassis has a ground clearance of $33\pm2~\mathrm{mm}$ and a $45^{\circ}$ inclined front deflector that redirects displaced water and mud away from the electronics. The same chassis also includes mounting points for the detachable flotation, camera, and microphone modules used in the outdoor demonstrations, while allowing additional sensing modules to be added in future deployments. The complete platform has a mass of $560~\mathrm{g}$, including approximately $40~\mathrm{g}$ per origami wheel, and is powered by a 3S, $850~\mathrm{mAh}$ lithium-polymer battery.

A Raspberry Pi Compute Module 4 running ROS~2 Humble commands the actuators at $50~\mathrm{Hz}$, interfaces with the sensors, and manages data logging. Encoder and actuator-current feedback from all actuators are recorded together with motion, voltage, and current measurements. During experiments, the robot is teleoperated wirelessly through onboard WiFi or Bluetooth. Visual, acoustic, and actuator data are logged onboard, and front-facing video is recorded at $30~\mathrm{fps}$.

\subsection{Kinematics and Odometry}
\label{subsec:kinematics}

\begin{figure}[hbt!]
    \centering
    \includegraphics[width=1\linewidth]{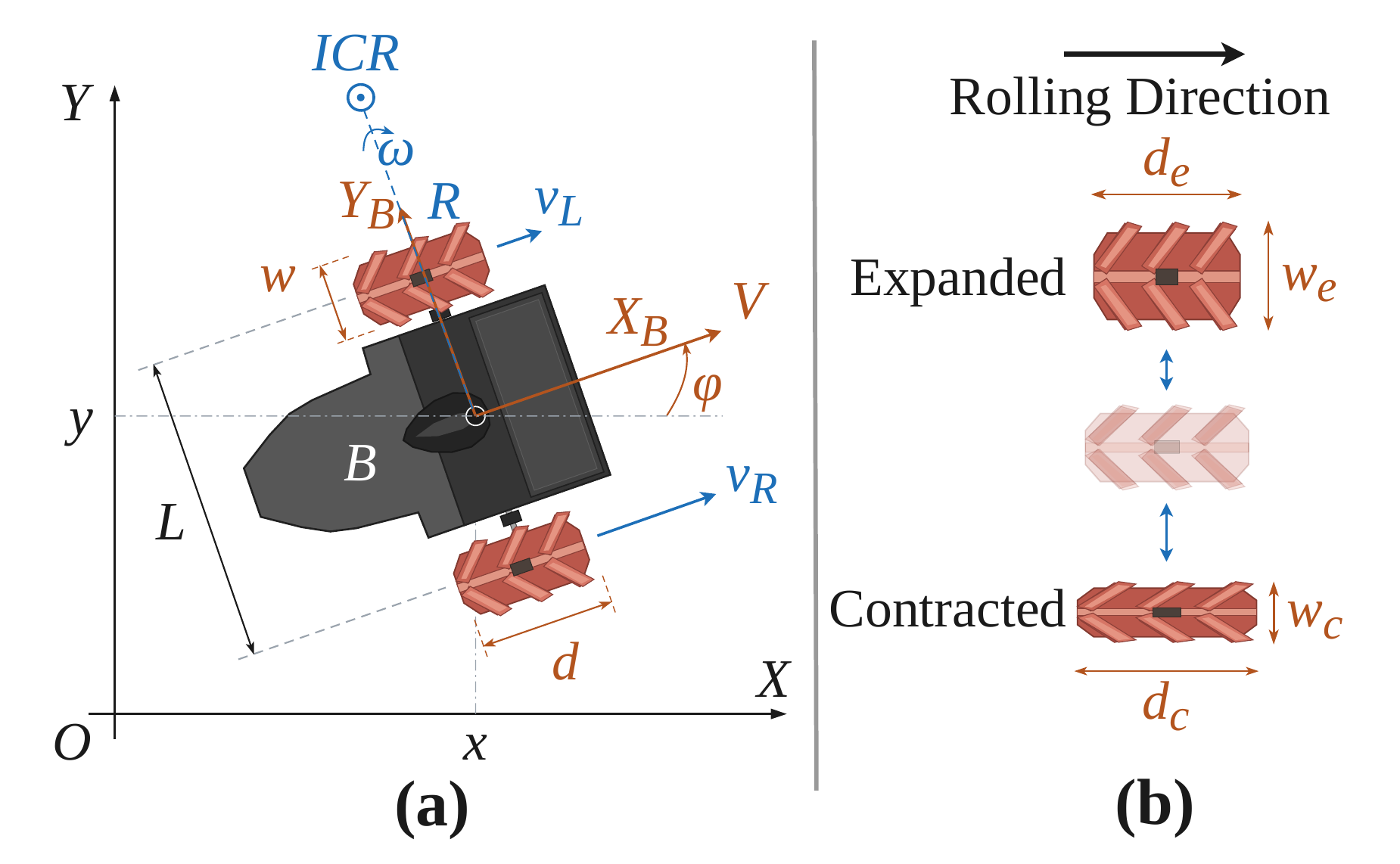}
    \caption{Differential-drive kinematics for the contracted and expanded wheel configurations. The velocity model uses the configuration-specific calibrated rolling radius, as defined in Sec.~\ref{subsec:mobility_metrics}.}
    \label{fig:kinematics}
\end{figure}

The robot was modelled using standard differential-drive kinematics. The model uses the configuration-specific calibrated rolling radius defined in Sec.~\ref{subsec:mobility_metrics}, rather than the geometric body radius. This distinction accounts for the effective contact radius of the complete wheel, including the fins. For the active wheel configuration $m\in\{\mathrm{C},\mathrm{E}\}$, the calibrated rolling radius is

\begin{equation}
R_{\mathrm{ref},m} =
\begin{cases}
R_{\mathrm{ref,C}}, & m=\mathrm{Contracted},\\
R_{\mathrm{ref,E}}, & m=\mathrm{Expanded}.
\end{cases}
\end{equation}

Given the measured signed angular velocities of the left and right wheels, $\omega_L$ and $\omega_R$, the robot linear velocity along its longitudinal axis and angular velocity about the vertical axis are

\begin{equation}
v=\frac{R_{\mathrm{ref},m}}{2}(\omega_R+\omega_L),
\qquad
\dot{\phi}=\frac{R_{\mathrm{ref},m}}{L}(\omega_R-\omega_L),
\end{equation}

where $L$ is the lateral distance between the centre lines of the left and right wheels. The active wheel configuration therefore determines the mapping from encoder speed to ground velocity through $R_{\mathrm{ref},m}$. The same model was used to calculate wheel-encoder odometry for the subsequent analysis.

\section{Experimental Protocol}
\label{sec:experiments}

\subsection{Controlled Terrains and Payloads}
\label{subsec:terrain_protocol}
Both wheel configurations were evaluated on the same robot under identical protocols, enabling a controlled comparison of their terrain-dependent mobility. Locomotion tests were conducted in an acrylic testbed measuring $100\times50\times10~\mathrm{cm}$, containing a $3~\mathrm{cm}$ layer of the selected substrate (Fig.~\ref{fig:multiterrain_results}b). For each preparation, $4.7~\mathrm{kg}$ of dry soil substrate was distributed and compacted in the testbed. Sand was tested as a contrasting granular terrain. Mud conditions were prepared by adding water to the dry soil on a dry-mass basis: water was added at 0.5, 1, or 2 times the initial dry-soil mass. These preparations were labelled $\mathrm{Mud0.5}$, $\mathrm{Mud1.0}$, and $\mathrm{Mud2.0}$, respectively. Each mixture was manually mixed for $10~\mathrm{min}$ and allowed to rest for $10~\mathrm{min}$ before testing. Two moisture sensors sampled separate locations to check preparation consistency. Their readings served only as a relative moisture check and were not used to define an independent terrain-strength scale.

The robot was tested without additional payload and with symmetric $100$- and $200~\mathrm{g}$ payloads placed near the wheel axle. Because the robot remained buoyant, water trials used only the unloaded and $100~\mathrm{g}$ conditions, with the latter representing a sensing payload. Additional payload would primarily increase immersion depth rather than wheel-ground loading, so the water test was not used to quantify a payload trend. All trials began with a fully charged battery. Wheel configuration was selected before each locomotion run.

\subsection{Kinematic Calibration and Mobility Metrics}
\label{subsec:mobility_metrics}

The reference rolling radius was calibrated separately for each wheel configuration on a low-slip rigid surface. A rigid wheel with geometric radius $46.0~\mathrm{mm}$ was first used as a calibration check. Three trials of ten wheel revolutions gave $\bar{R}_{\mathrm{rigid}}=44.7\pm0.24~\mathrm{mm}$, corresponding to a $2.8\%$ difference from the geometric radius. The morphing-wheel reference radii were then obtained using the same single-wheel procedure, giving $R_{\mathrm{ref,C}}=42.9~\mathrm{mm}$ and $R_{\mathrm{ref,E}}=49.0~\mathrm{mm}$ for the contracted and expanded configurations, respectively.

For each run, the non-slip reference velocity was calculated from the mean measured encoder speed,
\[
\bar{\omega}=\frac{\omega_R+\omega_L}{2},
\qquad
v_{\mathrm{ref},i}=\bar{\omega}R_{\mathrm{ref},i}.
\] Forward slip ratio was then calculated as
\begin{equation}
s_i=1-\frac{v_{\mathrm{actual},i}}{v_{\mathrm{ref},i}},
\label{eq:slip_rewrite}
\end{equation}
where $i\in\{\mathrm{C},\mathrm{E}\}$. The reference radii were kept fixed across all deformable-terrain tests. They were not recalculated using deformable terrain motion because doing so would incorporate terrain-induced slip into the reference itself.

Robot pose and translational motion were measured by Vicon, wheel motion by actuator encoders, and sinkage by side-view imaging. Hub sinkage was normalized by the configuration-specific calibrated rolling radius as \(S_{\mathrm{hub},i}=(h_{\mathrm{sink},i}/R_{\mathrm{ref},i})\times100\%\), where \(R_{\mathrm{ref,C}}=42.9~\mathrm{mm}\) and \(R_{\mathrm{ref,E}}=49.0~\mathrm{mm}\) for the contracted and expanded configurations, respectively. The primary terrestrial metrics were slip ratio, hub sinkage, mean robot speed, and locomotion energy per travelled metre. Water was treated as a separate locomotion mode. Only mean speed, locomotion energy per metre, and locomotion cost of transport were reported for water. Slip ratio and sinkage were not reported because fin-mediated paddle contact and buoyancy make the ground-based rolling-radius reference unsuitable for water propulsion. The locomotion cost of transport was calculated as
\begin{equation}
\mathrm{CoT}_{\mathrm{move}}=\frac{E_{\mathrm{move}}}{mgd},
\label{eq:cot_rewrite}
\end{equation}
where $E_{\mathrm{move}}$ is the locomotion energy measured while the robot travels over distance $d$. The electrical energy was obtained by integrating the synchronized voltage-current product over the locomotion interval, $E_{\mathrm{move}}=\int_{t_0}^{t_1}V(t)I(t)\,\mathrm{d}t$. This locomotion metric excludes both morphing-transient energy and shape-holding power, and therefore does not represent total mission energy.

\subsection{Transition and Data Handling}
\label{subsec:additional_protocols}

For the laboratory terrain-transition trials, the robot traversed the prepared soil-mud-water route with either the contracted or expanded wheel configuration held fixed throughout each run. No morphing command was issued during these trials, which compared transition performance between the contracted and expanded wheel configurations. Separate in-motion morphing trials changed the wheel configuration while the robot was moving, and no morphing failures were observed. Stationary morphing must overcome friction at the ground contact. During locomotion, wheel rotation helped reduce this opposing friction, making in-motion reconfiguration simpler in this design. Outdoor trials were used as qualitative demonstrations of operation on natural deformable terrain and through a terrestrial-aquatic transition. All primary prepared-terrain locomotion conditions were evaluated in five trials per terrain and wheel configuration. This included the unloaded, $100$ g, and $200$ g payload conditions. Group summaries are reported as medians and are interpreted descriptively rather than as statistical evidence.

\section{Results and Discussion}
\label{sec:results}

\subsection{Repeatable Morphing and Mechanical Operating Range}
\label{subsec:morphing_results}

In the separate morphing trials, the crank-driven mechanism successfully changed the wheels between the contracted and expanded configurations, taking $1.95\pm0.11~\mathrm{s}$ on average. The expanded-to-contracted transition consumed $2.26~\mathrm{J}$, whereas the contracted-to-expanded transition consumed $1.48~\mathrm{J}$. These results only describe the time and energy required for wheel morphing. They do not indicate that either wheel configuration is more energy-efficient over a complete mission.

In separate in-motion morphing trials, the wheel configuration was changed while the robot was moving and no morphing failures were observed. Stationary morphing must overcome friction at the ground contact. During locomotion, wheel rotation helped reduce this opposing friction, making in-motion morphing simpler in this design. These tests were separate from the fixed-mode soil-mud-water terrain-transition trials reported below.

\subsection{Terrain-Dependent Locomotion}
\label{subsec:terrain_results}

\begin{figure*}[htb!]
    \centering
    \includegraphics[width=1\linewidth]{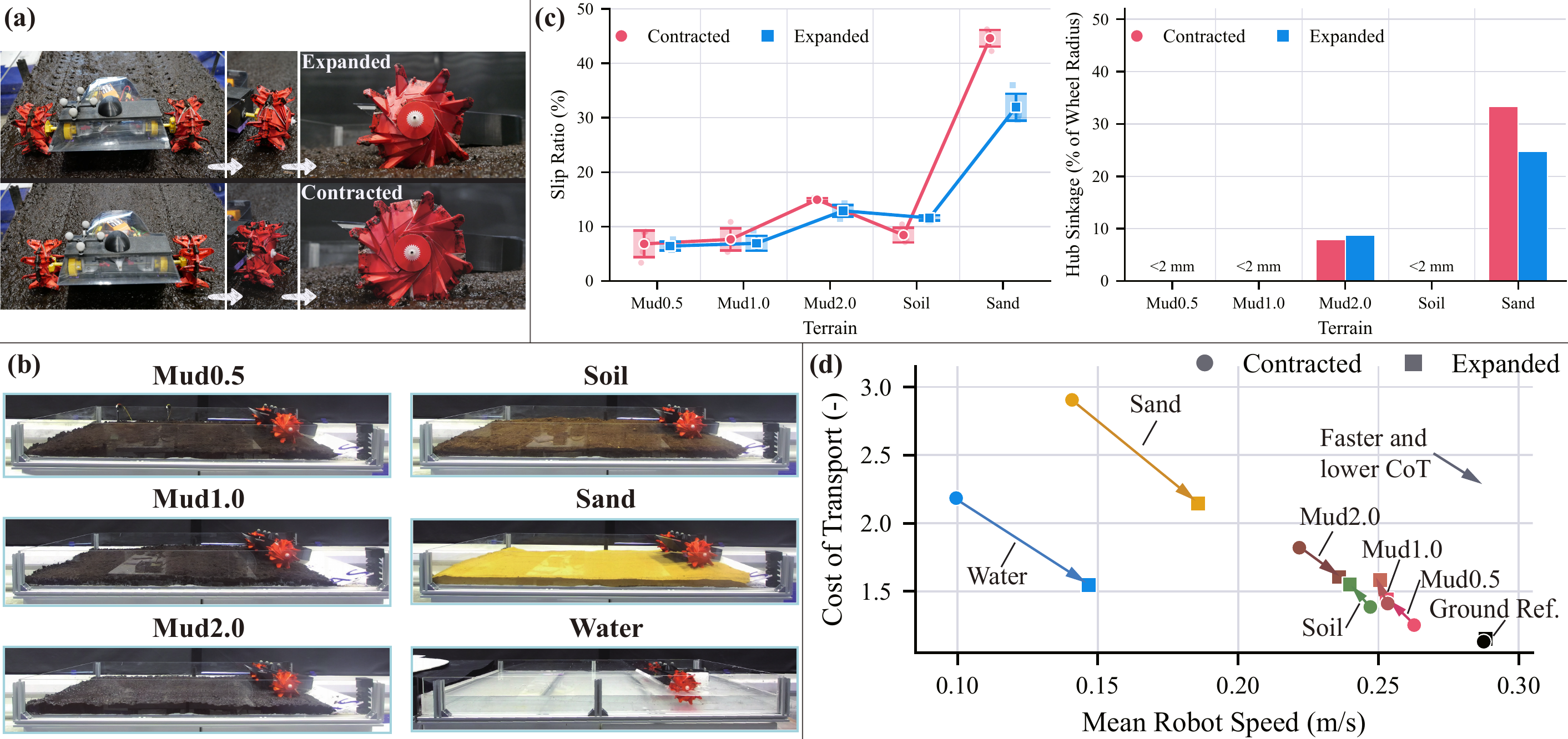} 
    \caption{Effects of terrain and wheel configuration on locomotion performance. (a) Representative images of MOSS after traversing Mud2.0 with the expanded and the contracted wheel configurations. The visible wheel imprints indicate the preceding motion and show the different wheel-terrain contact patterns. (b) Tested individual terrains. (c) Slip ratio and hub sinkage for various tested terrains. (d) cost of transport over mean robot speed. Each terrain-configuration combination was evaluated in five independent trials (n=5). Water is excluded from the terrestrial slip and hub-sinkage panels.}
    \label{fig:multiterrain_results}
\end{figure*}

Fig.~\ref{fig:multiterrain_results} provides qualitative context for the quantitative locomotion results. The distinct tracks on the mud surface show how the contact pattern changed between the two wheel configurations, while the photographs in Fig.~\ref{fig:multiterrain_results}b document the individual substrates used in the laboratory comparison.

The same robot and drive command were used to compare the contracted and expanded configurations across the prepared terrains (Fig.~\ref{fig:multiterrain_results}c). Neither configuration was uniformly preferable. Without a payload, the slip ratio remained below 8\% in Mud0.5 and Mud1.0, with only small differences between the two configurations. In Mud2.0, expansion reduced slip by about 2 percentage points, from 14.9\% to 12.9\%. The largest difference occurred in sand, where expansion reduced slip from approximately 45\% to 32\%, a decrease of about 13 percentage points. Soil showed the opposite trend, with lower slip for the contracted wheels, at 8.4\% compared with 11.6\% for the expanded wheels. Hub sinkage remained below the $2~\mathrm{mm}$ measurement resolution in Mud0.5, Mud1.0, and soil for both wheel configurations. In Mud2.0, the contracted wheel showed slightly less hub sinkage than the expanded wheel (7.9\% versus 8.8\%), despite greater slip. The expanded wheel's greater width and more inclined fins may have improved shear engagement while also causing greater local deformation in the weak, water-rich mud. Since wheel width, body diameter, and fin orientation change together, the difference cannot be attributed to any single feature. Sand produced the greatest sinkage, reaching 33.3\% for the contracted wheels compared with 24.7\% for the expanded wheels. Water was excluded from both the slip-ratio and sinkage analyses because buoyancy and fin-mediated paddle contact make the terrestrial ground-contact definitions unsuitable.

The CoT-speed plot in (Fig.~\ref{fig:multiterrain_results}d) summarizes the unloaded locomotion tests by placing the median locomotion cost of transport against the trial-level mean robot speed. The black marker shows the unloaded robot's mean speed and CoT measured on the low-slip rigid surface used for calibration. Arrows indicate the change from the contracted to the expanded configuration. Higher speed and lower CoT represent more favourable locomotion performance. Expansion of the wheel shifted the robot toward higher speed and lower CoT in Mud2.0, sand, and water. The largest terrestrial change occurred in sand, where speed increased from approximately 0.14 to 0.19 m/s while CoT decreased from about 2.9 to 2.1. By contrast, in Mud0.5, Mud1.0, and soil, the expanded configuration was slightly slower and had a higher CoT, so the contracted configuration offered the better speed-energy trade-off. The figure therefore shows that expansion was not universally advantageous. Its benefit became more obvious as the terrain became more deformable, particularly in Mud2.0 and sand, where the wider wheel configuration improved both speed and locomotion efficiency. In sand, the expanded configuration increased speed by approximately 32\% and reduced CoT by about 26\%. Water showed a similar shift, with speed increasing by approximately 47\% and CoT decreasing by about 29\%. Water was analyzed separately as an aquatic locomotion mode and is not directly comparable with terrestrial wheel-ground interaction. Because the analysis focuses on comparing locomotion performance, CoT is defined as locomotion energy per metre normalized by robot weight and excludes morphing transients and shape-holding power, so it does not represent total mission energy.

Overall, the results show that the preferable configuration depends on the terrain. The contracted wheel performed better on firmer, more supportive surfaces, whereas the expanded wheel most clearly reduced slip and sinkage in sand and reduced slip in Mud2.0. The wheel's larger terrain-contact envelope and reoriented fins likely improved engagement with deformable ground, but the coupled morphing motion prevents the separate effects of wheel width, body diameter, and fin orientation from being isolated.

\subsection{Payload Response}

\label{subsec:payload_results}
\begin{figure}[hbt!]
    \centering
    \includegraphics[width=\linewidth]{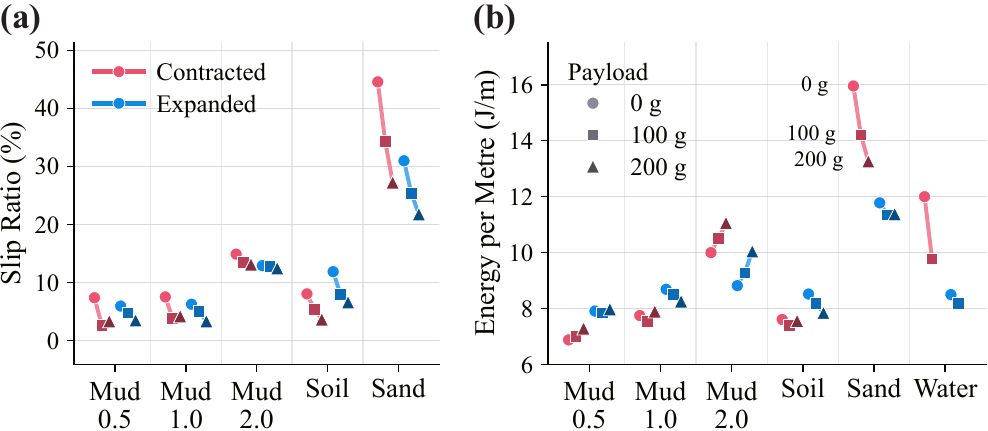} 
    
    \caption{Effects of payload on (a) slip ratio and (b) locomotion energy per metre across terrains and wheel configurations. Colours indicate wheel configuration, and marker shapes indicate payload. Markers and connecting lines show the median values for each terrain, configuration, and payload condition. Water is included only in panel (b).}
    \label{fig:results_load}
\end{figure}

The payload results (Fig.~\ref{fig:results_load}) show that added mass affected slip and energy differently across terrains. Added payload generally reduced slip on soil and sand, with the clearest reduction in sand, whereas the effect in Mud2.0 was small. This trend is consistent with the added normal load increasing the available shear force, although the benefit depended on the substrate. Locomotion energy per metre showed no universal payload trend. It increased with payload in Mud2.0 but changed little or decreased slightly in sand. The contracted configuration used less locomotion energy per metre in Mud0.5, Mud1.0, and soil, whereas the expanded configuration used less in Mud2.0 and sand. Water appears only in the energy panel and was analysed separately as an aquatic locomotion mode.

\begin{figure}[htb!]
    \centering
    \includegraphics[width=1\linewidth]{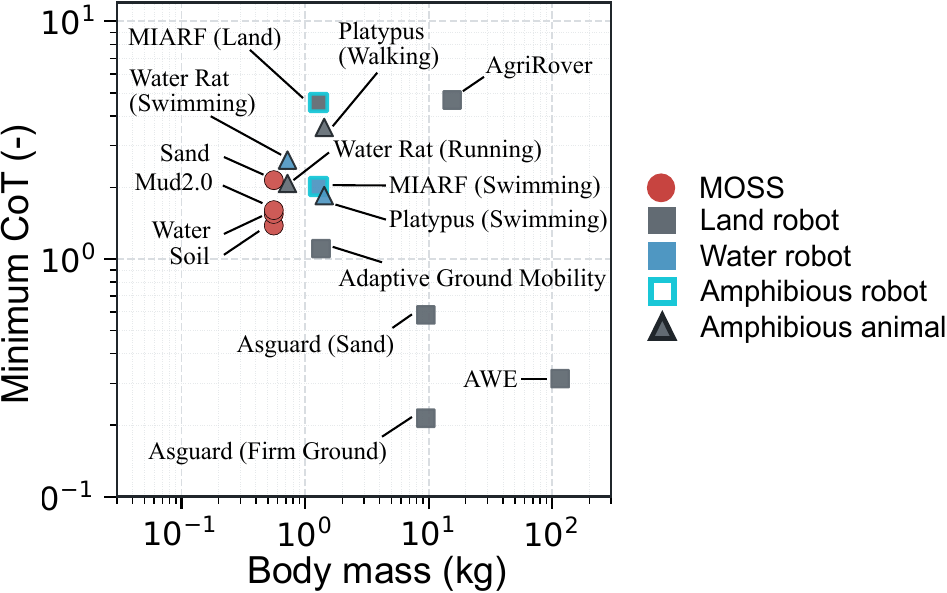}  
    \caption{Dimensionless cost of transport (CoT) versus body mass for MOSS, selected robots, and amphibious animals. Colours indicate terrestrial, aquatic, and amphibious locomotion. The comparison provides context relative to other selected systems. (AWE~\cite{papantoniou2020use}, Asguard~\cite{babu2010effects}, AgriRover~\cite{yan2020agrirover}, MIARF\cite{lin2023mudskipper}, and Adaptive Ground Mobility~\cite{kim2020using}, Australian water rat~\cite{fish1999energetics}, and Platypus~\cite{bethge2002energetics}.) }
    \label{fig:CoTcomparison}
\end{figure}

To place the locomotion energy requirements of MOSS in a broader context, Fig.~\ref{fig:CoTcomparison} compares its dimensionless CoT values with those of selected robotic systems and amphibians. Relative to MIARF and the amphibious animals, MOSS exhibits a lower or comparable CoT while supporting both aquatic and soft-terrain locomotion. Its deformable terrain performance also compares favourably with the agricultural rover included in the figure. Specialized terrestrial platforms such as Asguard and AWE achieve lower CoT, but their measurements correspond to dedicated ground locomotion on firmer or more uniform surfaces. Overall, MOSS combines moderate energy consumption with the ability to move across different swamp-relevant terrains, making it a versatile platform for challenging environments.

\subsection{Terrain-Transition Performance and Outdoor Demonstrations}

\begin{figure}[htb!]
    \centering
    \includegraphics[width=1\linewidth]{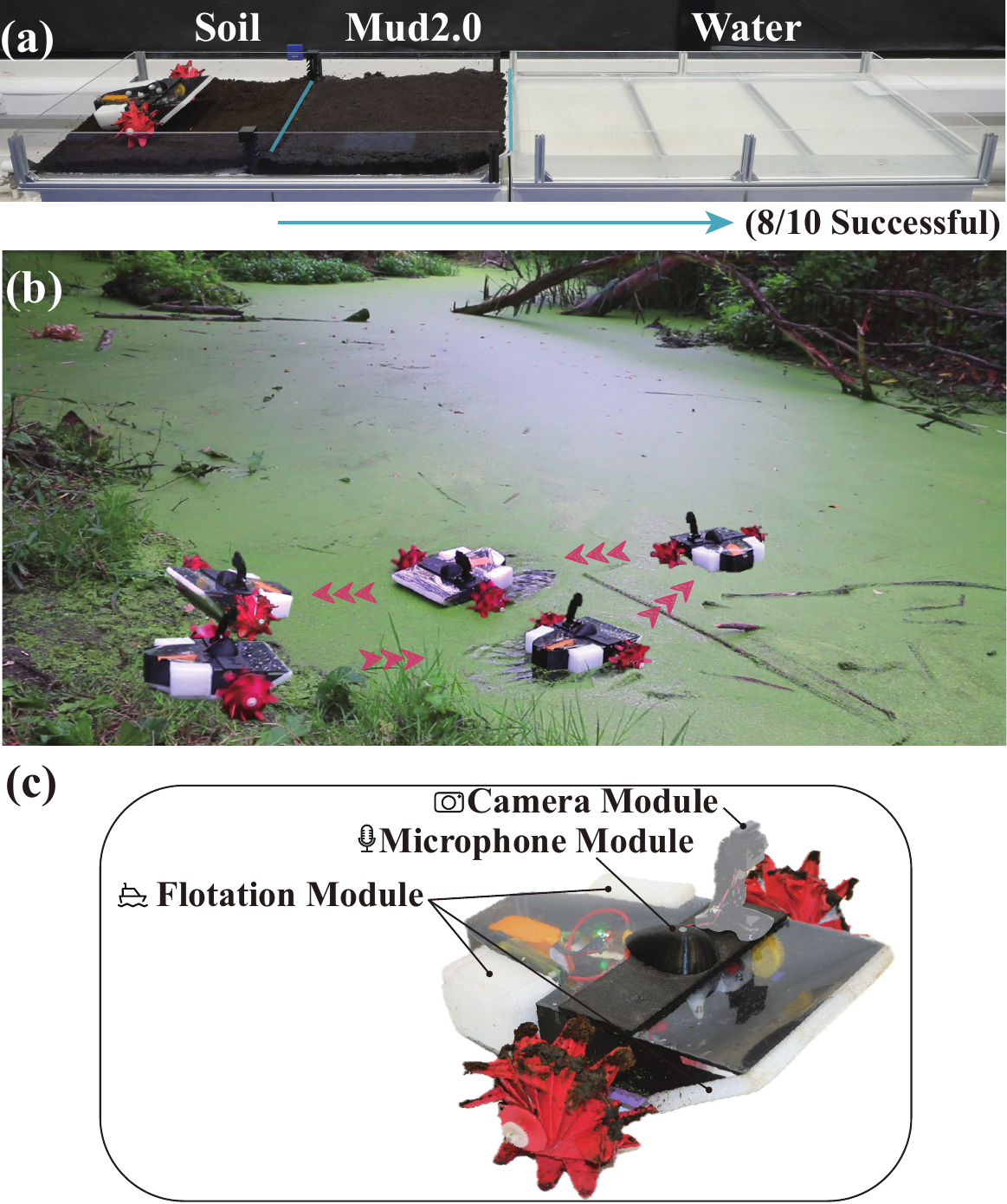}  
    \caption{Terrain transition and outdoor demonstration. (a) Connected testbed for the controlled soil-mud-water transition. 8 of 10 traversals succeeded. The transition test used one fixed wheel configuration per run, with no morphing command issued during the traversal. (b) Image sequence showing the outdoor transition test. (c) MOSS with detachable sensing modules mounted for field operation.}
    \label{fig:outdoor}
\end{figure}

In the laboratory transition test (Fig.~\ref{fig:outdoor}a), the robot traversed a prepared soil-mud-water route with either wheel configuration held throughout each run. Eight of ten traversals succeeded, and both failures occurred when the chassis contacted the rigid mud-water boundary, indicating a chassis-clearance limitation at this interface.

During outdoor field trials (Fig.~\ref{fig:outdoor}b), the robot traversed complex natural deformable terrain, operated in water, and crossed a terrestrial-aquatic transition while equipped with detachable flotation, camera, and microphone modules. These trials qualitatively demonstrated operation in natural conditions..

MOSS uses a modular payload interface for environment-specific hardware. The chassis provides mechanical mounting points, and the CM4 carrier board provides electrical interfaces for interchangeable modules. The interface allows different sensing payloads to be exchanged without modifying the main locomotion system. In the field demonstrations, detachable foam flotation, camera, and microphone modules were installed (Fig.~\ref{fig:outdoor}c). The flotation supported water operation, while the expanded wheel presented a larger submerged volume and therefore contributed additional buoyancy by displacing more water. The fins generated low-speed paddle-like thrust. The camera and microphone illustrate potential applications in ecosystem monitoring, including visual habitat surveys and acoustic data collection. These signals were recorded for demonstration only and were not analysed in this study.

\section{Conclusion}
\label{sec:conclusion}
This work presented MOSS, a compact amphibious robot designed for locomotion across swamp-relevant deformable terrain. Its 3D-printed origami wheels change their width, body diameter, and fin orientation through a single morphing input. A telescopic shaft separates axial morphing from torque transmission and provides a guided load path. The more energy-efficient configuration varied by terrain. The contracted configuration required less energy per metre in soil, Mud0.5, and Mud1.0, whereas the expanded configuration required less in Mud2.0, sand, and water. Expanded wheels also reduced slip and hub sinkage in sand and increased speed in water. These results indicate that morphing provides two wheel–surface interaction modes. The contracted configuration favours efficient rolling on more supportive surfaces, whereas the expanded configuration provides greater contact and terrain engagement when slip or sinkage becomes limiting. Separate in-motion morphing trials produced no observed failures. In the soil–mud–water transition test, 8 of 10 traversals succeeded, with both failures caused by limited chassis clearance at the rigid mud–water boundary. Comparative CoT analysis showed that MOSS achieves competitive locomotion efficiency while maintaining mobility across aquatic and deformable terrain environments. Outdoor field trials further demonstrated operation on natural deformable
terrain, in water, and across a terrestrial-aquatic transition with modular
sensing payloads. Overall, these results demonstrate that wheel morphing can provide terrain-dependent locomotion benefits and support operation across swamp-relevant deformable terrain, water, and terrestrial-aquatic transitions.

Future work will improve traction with higher-friction contact materials and reduce shape-holding power with a self-locking morphing mechanism. The modular interface could support microphones, environmental sampling tools, and infrared cameras for biodiversity monitoring and post-flood or mudslide inspection. Future versions will combine onboard terrain sensing with autonomous navigation to select the most suitable wheel configuration during operation. Broader field trials will then evaluate this terrain-aware control strategy.





 \bibliographystyle{IEEEtran}

 \bibliography{mybib}

@article{lough2026rolling,
  title={Rolling into a ball (volvation) provides an antipredatory benefit},
  author={Lough, Mason W and Emberts, Zachary},
  journal={Animal Behaviour},
  volume={236},
  pages={123578},
  year={2026},
  publisher={Elsevier}
}

@article{callens2018flat,
  title={From flat sheets to curved geometries: Origami and kirigami approaches},
  author={Callens, Sebastien JP and Zadpoor, Amir A},
  journal={Materials Today},
  volume={21},
  number={3},
  pages={241--264},
  year={2018},
  publisher={Elsevier}
}

@inproceedings{rhoads2016design,
  title={The design and fabrication of a deformable origami wheel},
  author={Rhoads, Benjamin P and Su, Hai-Jun},
  booktitle={International Design Engineering Technical Conferences and Computers and Information in Engineering Conference},
  volume={50169},
  pages={V05BT07A021},
  year={2016},
  organization={American Society of Mechanical Engineers}
}

@article{chen2015origami,
  title={Origami of thick panels},
  author={Chen, Yan and Peng, Rui and You, Zhong},
  journal={Science},
  volume={349},
  number={6246},
  pages={396--400},
  year={2015},
  publisher={American Association for the Advancement of Science}
}

@inproceedings{zhou2017multimodal,
  title={A multimodal hybrid robot with transformable wheels},
  author={Zhou, Faliang and others},
  booktitle={2017 IEEE International Conference on Real-time Computing and Robotics (RCAR)},
  pages={139--144},
  year={2017},
  organization={IEEE}
}

@article{pace2009mudskipper,
  title={Mudskipper pectoral fin kinematics in aquatic and terrestrial environments},
  author={Pace, CM and Gibb, AC},
  journal={Journal of Experimental Biology},
  volume={212},
  number={14},
  pages={2279--2286},
  year={2009},
  publisher={Company of Biologists}
}

@article{dunbabin2012robots,
  title={Robots for environmental monitoring: Significant advancements and applications},
  author={Dunbabin, Matthew and Marques, Lino},
  journal={IEEE Robotics \& Automation Magazine},
  volume={19},
  number={1},
  pages={24--39},
  year={2012},
  publisher={IEEE}
}

@article{lowman2021life,
  title={Life in the treetops—an overview of forest canopy science and its future directions},
  author={Lowman, Margaret},
  journal={Plants, People, Planet},
  volume={3},
  number={1},
  pages={16--21},
  year={2021},
  publisher={Wiley Online Library}
}

@article{bujari2020mobile,
  title={A mobile sensing and visualization platform for environmental data},
  author={Bujari, Armir and Gaggi, Ombretta and Palazzi, Claudio E},
  journal={Pervasive and Mobile Computing},
  volume={66},
  pages={101204},
  year={2020},
  publisher={Elsevier}
}

@article{pimm2014biodiversity,
  title={The biodiversity of species and their rates of extinction, distribution, and protection},
  author={Pimm, Stuart L and others},
  journal={science},
  volume={344},
  number={6187},
  pages={1246752},
  year={2014},
  publisher={American Association for the Advancement of Science}
}

@book{reid2005ecosystems,
  title={Ecosystems and human well-being-Synthesis: A report of the Millennium Ecosystem Assessment},
  author={Reid, Walter V and Mooney, Harold A and Cropper, Angela and Capistrano, Doris and Carpenter, Stephen R and Chopra, Kanchan and Dasgupta, Partha and Dietz, Thomas and Duraiappah, Anantha Kumar and Hassan, Rashid and others},
  year={2005},
  publisher={Island Press}
}

@article{siddall2014launching,
  title={Launching the AquaMAV: bioinspired design for aerial--aquatic robotic platforms},
  author={Siddall, R and Kova{\v{c}}, M},
  journal={Bioinspiration \& biomimetics},
  volume={9},
  number={3},
  pages={031001},
  year={2014},
  publisher={IOP Publishing}
}

@article{debruyn2020medusa,
  title={Medusa: A multi-environment dual-robot for underwater sample acquisition},
  author={Debruyn, Diego and Zufferey, Raphael and Armanini, Sophie F and Winston, Crystal and Farinha, Andr{\'e} and Jin, Yufei and Kovac, Mirko},
  journal={IEEE Robotics and Automation Letters},
  volume={5},
  number={3},
  pages={4564--4571},
  year={2020},
  publisher={IEEE}
}

@inproceedings{lee2013deformable,
  title={The deformable wheel robot using magic-ball origami structure},
  author={Lee, Dae-Young and Kim, Ji-Suk and Kim, Sa-Reum and Koh, Je-Sung and Cho, Kyu-Jin},
  booktitle={International design engineering technical conferences and computers and information in engineering conference},
  volume={55942},
  pages={V06BT07A040},
  year={2013},
  organization={American Society of Mechanical Engineers}
}

@misc{worldrainforests_swamps,
  author       = {{World Rainforests}},
  title        = {Rainforest Rivers, Lakes and Swamps},
  year         = {n.d.},

  note         = {Accessed: 2026-05-30}
}

@article{lee2017origami,
  title={Origami wheel transformer: A variable-diameter wheel drive robot using an origami structure},
  author={Lee, Dae-Young and Kim, Sa-Reum and Kim, Ji-Suk and Park, Jae-Jun and Cho, Kyu-Jin},
  journal={Soft robotics},
  volume={4},
  number={2},
  pages={163--180},
  year={2017},
  publisher={SAGE Publications Sage CA: Los Angeles, CA}
}

@article{moger_design_2024,
  title={Design and implementation of a mobile robot with variable-diameter wheels},
  author={Moger, Gourav and Varol, Huseyin Atakan},
  journal={IEEE/ASME Transactions on Mechatronics},
  volume={30},
  number={4},
  pages={2529--2538},
  year={2024},
  publisher={IEEE}
}

@article{liu_origami-wheeled_2025,
	title = {An origami-wheeled robot with variable width and enhanced sand walking versatility},
	volume = {206},
	doi = {10.1016/j.tws.2024.112645},
	journal = {Thin-Walled Structures},
	author = {Liu, Jie and Pang, Zufeng and Li, Zhiyong and Wen, Guilin and Su, Zhoucheng and He, Junfeng and Liu, Kaiyue and Jiang, Dezheng and Li, Zenan and Chen, Shouyan and Tian, Yang and Xie, Yi and Wang, Zhen-Pei and Liu, Zhuangjian},
	month = jan,
	year = {2025},
	pages = {112645},
}

@article{mitani_design_2009,
  title={A design method for 3D origami based on rotational sweep},
  author={Mitani, Jun},
  journal={Computer-Aided Design and Applications},
  volume={6},
  number={1},
  pages={69--79},
  year={2009},
  publisher={Taylor \& Francis}
}

@inproceedings{murphy2024evaluation,
  title={Evaluation and design recommendations for a folding morphing-wheg robot for nuclear characterisation},
  author={Murphy, Dominic and Giuliani, Manuel and Bremner, Paul},
  booktitle={2024 IEEE/RSJ International Conference on Intelligent Robots and Systems (IROS)},
  pages={9088--9093},
  year={2024},
  organization={IEEE}
}

@misc{easycomposites_carbon_fibre_pultrusions_2024,
  author       = {{Easy Composites Ltd.}},
  title        = {Carbon Fibre Pultrusions},
  year         = {2024},
  note         = {Technical datasheet, Version 1.0, revised 20 September 2024},
  ignoredurl         = {https://media.easycomposites.co.uk/datasheets/EC-TDS-Carbon-Fibre-Pultrusions.pdf},
  urldate      = {2026-08-10}
}

@inproceedings{robertson2025global,
  title={Global Wetland Outlook 2025: Valuing, conserving, restoring and financing wetlands},
  author={Robertson, Hugh A},
  year={2025},
  organization={Ramsar Convention on Wetlands}
}

@article{fish1999energetics,
  title={Energetics of locomotion by the Australian water rat (Hydromys chrysogaster): a comparison of swimming and running in a semi-aquatic mammal},
  author={Fish, Frank E and Baudinette, Russell Victor},
  journal={Journal of Experimental Biology},
  volume={202},
  number={4},
  pages={353--363},
  year={1999},
  publisher={The Company of Biologists Ltd}
}

@book{bethge2002energetics,
  title={Energetics and foraging behaviour of the platypus, Ornithorhynchus anatinus},
  author={Bethge, Philip},
  year={2002},
  publisher={University of Tasmania Hobart, Tasmania, Australia}
}

@inproceedings{papantoniou2020use,
  title={Use of flexible variable stiffness wheels for high speed lunar and planetary exploration rovers and their potential impact on the cost of transport},
  author={Papantoniou, Vassilios and others},
  booktitle={Proc. of the I-SAIRAS},
  pages={18--21},
  year={2020}
}

@inproceedings{babu2010effects,
  title={Effects of wheel synchronization for the hybrid leg--wheel robot asguard},
  author={Babu, Ajish and others},
  booktitle={Proceedings of International Symposium on Artificial Intelligence, Robotics and Automation in Space (iSAIRAS-10), Sapporo},
  year={2010}
}

@incollection{yan2020agrirover,
  title={The agrirover: a reinvented mechatronic platform from space robotics for precision farming},
  author={Yan, Xiu-Tian and others},
  booktitle={Reinventing mechatronics: Developing future directions for mechatronics},
  pages={55--73},
  year={2020},
  publisher={Springer}
}

@article{kim2014wheel,
  title={Wheel transformer: A wheel-leg hybrid robot with passive transformable wheels},
  author={Kim, Yoo-Seok and others},
  journal={IEEE Transactions on Robotics},
  volume={30},
  number={6},
  pages={1487--1498},
  year={2014},
  publisher={IEEE}
}

@article{lin2023mudskipper,
  title={Mudskipper-inspired amphibious robotic fish enhances locomotion performance by pectoral-caudal fins coordination},
  author={Lin, Zhonglu and others},
  journal={Cell Reports Physical Science},
  volume={4},
  number={10},
  year={2023},
  publisher={Elsevier}
}

@inproceedings{kim2020using,
  title={Using manipulation to enable adaptive ground mobility},
  author={Kim, Raymond and others},
  booktitle={2020 IEEE International Conference on Robotics and Automation (ICRA)},
  pages={857--863},
  year={2020},
  organization={IEEE}
}

@article{lee2024variable,
  title={Variable-stiffness--morphing wheel inspired by the surface tension of a liquid droplet},
  author={Lee, Jae-Young and others},
  journal={Science Robotics},
  volume={9},
  number={93},
  pages={eadl2067},
  year={2024},
  publisher={American Association for the Advancement of Science}
}

\vspace{12pt}
\color{red}

\end{document}